\documentclass[11pt]{article}

\usepackage{iftex}
\usepackage{subcaption}
\usepackage{graphicx}

\usepackage{booktabs}
\usepackage{multirow}
\usepackage{tabularx}

\usepackage{acl}

\usepackage{times}
\usepackage{latexsym}

\usepackage[T1]{fontenc}

\usepackage[utf8]{inputenc}

\usepackage{microtype}

\usepackage{inconsolata}

\usepackage{multirow}
\usepackage{graphicx}
\usepackage{amsmath}
\usepackage{svg}
\usepackage{amsfonts}

\usepackage[arabic,english]{babel}

\title{From FusHa to Folk: Exploring Cross-Lingual Transfer in Arabic Language Models}

\author{{\bf Abdulmuizz Khalak} \qquad {\bf Abderrahmane Issam} \qquad {\bf Gerasimos Spanakis} \\
        Department of Advanced Computing Sciences \\ 
        Maastricht University \\ 
        \small{\texttt{\{a.khalak@student., abderrahmane.issam@, jerry.spanakis@\}maastrichtuniversity.nl}}}

\begin{document}
\maketitle

\begin{abstract}

Arabic Language Models (LMs) are pretrained predominately on Modern Standard Arabic (MSA) and are expected to transfer to its dialects. While MSA as the standard written variety is commonly used in formal settings, people speak and write online in various dialects that are spread across the Arab region. This poses limitations for Arabic LMs, since its dialects vary in their similarity to MSA. In this work we study cross-lingual transfer of Arabic models using probing on 3 Natural Language Processing (NLP) Tasks, and representational similarity. Our results indicate that transfer is possible but disproportionate across dialects, which we find to be partially explained by their geographic proximity. Furthermore, we find evidence for negative interference in models trained to support all Arabic dialects. This questions their degree of similarity, and raises concerns for cross-lingual transfer in Arabic models. \footnote{\scriptsize Our code: \texttt{\url{https://github.com/muizzkhalak/cross_lingual_transfer_arabic}}}

\end{abstract}

\section{Introduction} \label{introduction}
Arabic is a major world language with over 400 million speakers and a central role in cultural, historical, and religious life across the Middle East and North Africa (MENA) \cite{owens2006, versteegh2014}. Structurally, it is a \textit{Semitic} language with a long written tradition, but for NLP the most salient property is its pervasive \textit{diglossia}: a standardized written variety, Modern Standard Arabic (MSA), coexists with a rich spectrum of non-standardized dialects that vary across the MENA region \cite{ferguson1959}. MSA (FusHa) dominates education, news, and formal writing, whereas dialects are the default medium of everyday communication and online interaction.

Although Modern Standard Arabic (MSA) is far less frequent in daily speech than Dialectal Arabic (DA), it remains the dominant variety in the digital corpora used to train Large Language Models (LLMs). Consequently, most existing Arabic LMs are predominantly MSA-centric, with limited and disproportionate coverage of diverse dialects \cite{antoun-etal-2020-arabert,inoue2021interplayvariantsizetask,abdul-mageed-etal-2021-arbert,antoun-etal-2021-aragpt2,sengupta2023jaisjaischatarabiccentricfoundation,lan-etal-2020-empirical}. While these models are often implicitly expected to generalize across the Arabic continuum, cross-lingual transfer is not guaranteed, particularly as many dialects exhibit low mutual intelligibility with MSA and one another \cite{abu-farha-magdy-2022-effect, keleg-etal-2023-aldi}. This linguistic imbalance has spurred the development of dialect-specific models \cite{qarah2024saudibertlargelanguagemodel,gaanoun2023darijabert,abdaoui2022dziribertpretrainedlanguagemodel,qarah2024egybertlargelanguagemodel,shang2025nilechategyptianlanguagemodels,alyami-al-zaidy-2022-weakly}. However, the necessity of such specialization remains an open question: \textbf{Do MSA-centric or multi-dialect models transfer equitably across the Arabic dialectal landscape?} To investigate this, we probe the internal representations of Arabic LMs to evaluate their transferability to various dialects across three core NLP tasks—Sentiment Analysis (SA), Named Entity Recognition (NER), and Part-of-Speech (POS) tagging—benchmarking their performance against specialized, dialect-specific counterparts.

While probing provides a functional account of cross-lingual transfer, its scope is often constrained by specific task selections and potential confounding dataset characteristics. Prior research has leveraged annotated linguistic features to predict inter-lingual transfer or interference \cite{lin-etal-2019-choosing, ERONEN2023103250}; however, the scarcity of granular linguistic annotations for Arabic dialects renders such approaches currently unfeasible. While \citet{ALSUDAIS2022102770} utilized lexical overlap within the parallel MADAR corpus \cite{bouamor-etal-2018-madar} as a proxy for dialectal similarity, surface-level overlap fails to adequately capture deeper syntactic and semantic nuances. Given the proficiency of LMs in encoding these high-level features \cite{conneau-etal-2018-cram}, we propose employing Representational Similarity Analysis (RSA) between MSA and DA model representations. Specifically, we utilize Centered Kernel Alignment (CKA) \cite{pmlr-v97-kornblith19a} to quantify layer-wise similarity between DA and MSA models using parallel sentences from the MADAR dataset. We posit that this provides an intrinsic metric of cross-lingual transfer: high representational similarity between MSA and DA models serves as a robust signal for effective transferability and suggests that inter-dialectal interference will be minimized.

We integrate these two methodologies---Probing and Representational Similarity Analysis (RSA)---to provide a holistic evaluation of cross-lingual transfer between MSA and its dialects. Our findings indicate that, in general, MSA-centric models demonstrate strong transferability to DA, occasionally outperforming dialect-specific models. However, this transfer remains significantly disproportionate across the dialectal spectrum. To investigate the drivers of this disparity, we correlate model performance with factors such as geographic proximity and pretraining data volume. Our analysis reveals that dialect-specific models consistently exceed the performance of MSA and multi-dialect models only when supported by substantial DA pretraining data. Furthermore, geographic proximity serves as a strong predictor of transferability, aligning with the dialectal continuum hypothesis: as geographic distance increases, mutual intelligibility diminishes, posing a greater challenge for MSA-to-DA transfer.

\section{Related Work} \label{related_work}

\paragraph{Cross-Lingual Transfer.}
Prior work consistently shows that transfer effectiveness depends strongly on source--target similarity: typologically or linguistically closer languages yield better zero-shot transfer, and language-distance metrics correlate with downstream accuracy \cite{ERONEN2023103250,philippy-etal-2023-identifying}. Consequently, selecting a related transfer source—rather than defaulting to English—can substantially improve performance for low-resource targets \cite{lin-etal-2019-choosing,ERONEN2023103250}. Multilingual LMs such as mBERT further demonstrate robust cross-lingual generalization without explicit alignment, achieving competitive zero-shot results across tasks like Natural Language Inference (NLI) and NER after fine-tuning on a single language \cite{conneau-etal-2018-xnli,pires-etal-2019-multilingual,wu-dredze-2019-beto}. This behavior is often attributed to increasingly language-agnostic representations in higher layers \cite{k2020crosslingualabilitymultilingualbert,artetxe-etal-2020-cross}; notably, retraining only the embedding layer of a monolingual transformer can effectively adapt it to a new language, suggesting that upper-layer features are largely transferable \cite{artetxe-etal-2020-cross}. However, multilingual training often leads to negative interference especially for high resource languages \cite{bapna-firat-2019-simple, chang2023multilingualitycurselanguagemodeling,alastruey2025interferencematrixquantifyingcrosslingual}. Overall, these findings motivate balanced transfer strategies, where leveraging one or a few related high-resource languages yields strong gains for low-resource targets without incurring excessive interference \cite{seto-etal-2025-training}. In this study, we examine how well MSA-centric and multi-dialect Arabic models transfer to DA varieties, and whether negative interference occurs in dialectal context, even when linguistic similarity is assumed. Although our process follows cross-dialectal transfer across Arabic models, we use "cross-lingual" in the rest of the paper to align with literature norms.

\paragraph{Probing Arabic Language Models.}
Previous work has examined how linguistic information is distributed across layers and neurons in Arabic transformer models, including dialect-trained variants \cite{abdelali-etal-2022-post}. This line of research reports that lower and middle layers primarily encode morphology, upper layers capture syntactic dependencies, and MSA-based models often fail to represent dialect-specific nuances despite substantial vocabulary overlap; embedding-layer neurons tend to be polysemous, whereas mid-layer neurons are more specialized for particular linguistic properties \cite{abdelali-etal-2022-post}. By contrast, we focus on model-level representations: we assess how \textbf{MSA}, general-purpose Arabic (\textbf{MIX}/\textbf{Multi-DA}), and dialect-specific encoders encode morphological and syntactic information via controlled probing tasks, comparing best-layer performance across models rather than inspecting individual neurons.

\paragraph{Similarity of Arabic Dialects.}

Recent work in Arabic dialectometry links textual similarity to geography, showing that geographically proximate varieties tend to be linguistically closer, sometimes more so than political boundaries \cite{ALSUDAIS2022102770}. Across dialectal corpora, classic NLP/IR similarity measures further reveal strong Levantine cohesion and suggest that Palestinian is particularly close to MSA \cite{KWAIK20182}, while comparisons with Classical Arabic provide a historical reference for modern variation \cite{abouzahr2025lexical}. Analyses of the MADAR corpus also report Muscat (Spoken in Oman) as among the closest dialects to MSA and Sfax (Spoken in Tunisia) among the most distant \cite{bouamor-etal-2018-madar}. These lexical analyses remain limited at capturing syntactic and semantic similarity, therefore, our work studies whether such proximity relationships are internalized by pre-trained language models, which were shown to capture syntactic and semantic features.

\section{Methodology} \label{methodology}

In this section, we outline our methodological framework for studying cross-lingual transfer of Arabic LMs. We combine two complementary methods for this study, namely: linear probes trained on frozen layer-wise embeddings (Section~\ref{probing}), and similarity of model representations measured using Centered Kernel Alignment (CKA) (Section~\ref{cka}). Furthermore, motivated by the linguistic theory of dialect continuum and its implications \cite{Chambers_Trudgill_1998}, we propose a proxy for measuring geographic proximity to MSA (Section~\ref{sec:geo_continuum}).

\subsection{Probing Classifiers} \label{probing}

Probing has been widely applied to analyze multilingual models, where it can reveal whether models encode language-specific features or universal cross-linguistic structures \cite{DBLP:journals/corr/abs-1905-06316}. In the case of Arabic, probing classifiers can help determine whether dialectal and MSA models capture similar linguistic phenomena and whether knowledge acquired in one variety (e.g., MSA) transfers to another (e.g., Egyptian or Levantine Arabic). By training probes on embeddings derived from dataset of one variety using a language model of different variety. 

\begin{figure}[tbp]
    \centering
    \includegraphics[width=0.7\columnwidth]{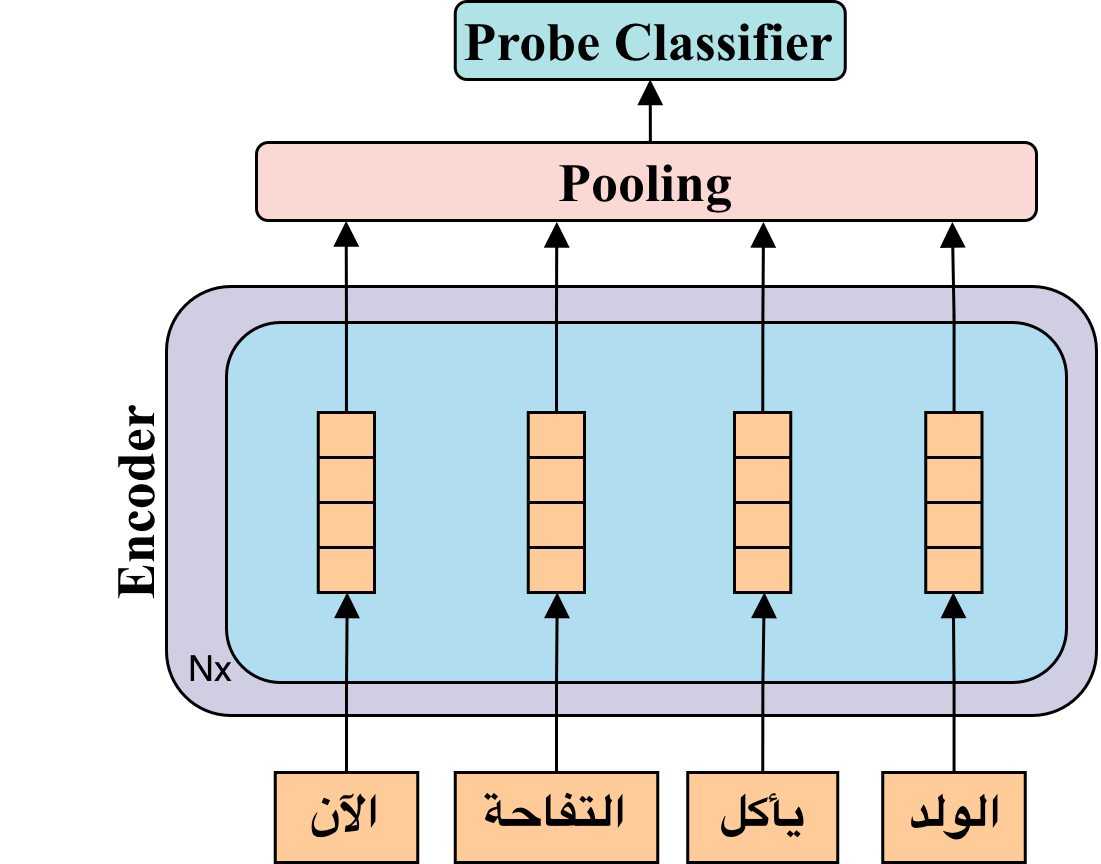}
    \caption{Architecture of the probing classifier for the example sentence “The boy is eating the apple now.” Sentence representations pass through N layers, and each layer is probed using the classifier in Eq.~\ref{eq:Probe}.}
    \label{fig:Probing_Classifier}
\end{figure}

The standard probing setup involves training a lightweight classifier—commonly a linear or shallow feedforward model—on top of frozen representations extracted from different layers of the LM. We train probes to predict a supervised linguistic property (e.g., POS tags) using embeddings as input. If the probe achieves high accuracy, this suggests that the relevant linguistic information is encoded in the representations at that layer and is easily extractable by a simple function \cite{hewitt-manning-2019-structural}. Probing classifiers are deliberately kept simple to minimize the chance that the probe itself learns the task from scratch. Instead, their purpose is to act as a diagnostic tool that reveals the presence (or absence) of linguistic features in the LM representations \cite{pimentel-etal-2020-information}. Figure \ref{fig:Probing_Classifier} illustrates the probing architecture, where we employ a linear classifier with a softmax activation layer. For a given layer $l$, the probability distribution $y_l$ is computed as:

\begin{equation} \label{eq:Probe}
    y_l = \mathrm{softmax}(W \mathbf{z}_l + b)
\end{equation}
where $\mathbf{z}_l$ is a representation derived from the hidden states $H_l = \{h_{l,1}, \dots, h_{l,T}\}$. We define two pooling strategies depending on the probing task:
\begin{itemize}
    \item \textbf{Token-level} to POS and NER tasks: $\mathbf{z}_l = h_{l,i}$, the embedding at index $i$.
    \item \textbf{Sentence-level} for the SA task: $\mathbf{z}_l = \frac{1}{T} \sum_{i=1}^{T} h_{k,i}$, the mean-pooled average of all token embeddings in the sequence.
\end{itemize}

\subsection{Representation Similarity Analysis} \label{cka}

Besides cross-lingual transfer as measured using probing on selected tasks, we also compute representation similarity, which can be viewed as an intrinsic signal for transfer. Primarily, we seek to study whether MSA LMs learn similar representations to their DA counterpart, which can be seen as a signal of similarity between DA varieties and MSA, and of the sufficiency of MSA for capturing the nuances of DA varieties.

\begin{figure}[tbp]
    \centering
    \includegraphics[width=0.9\columnwidth]{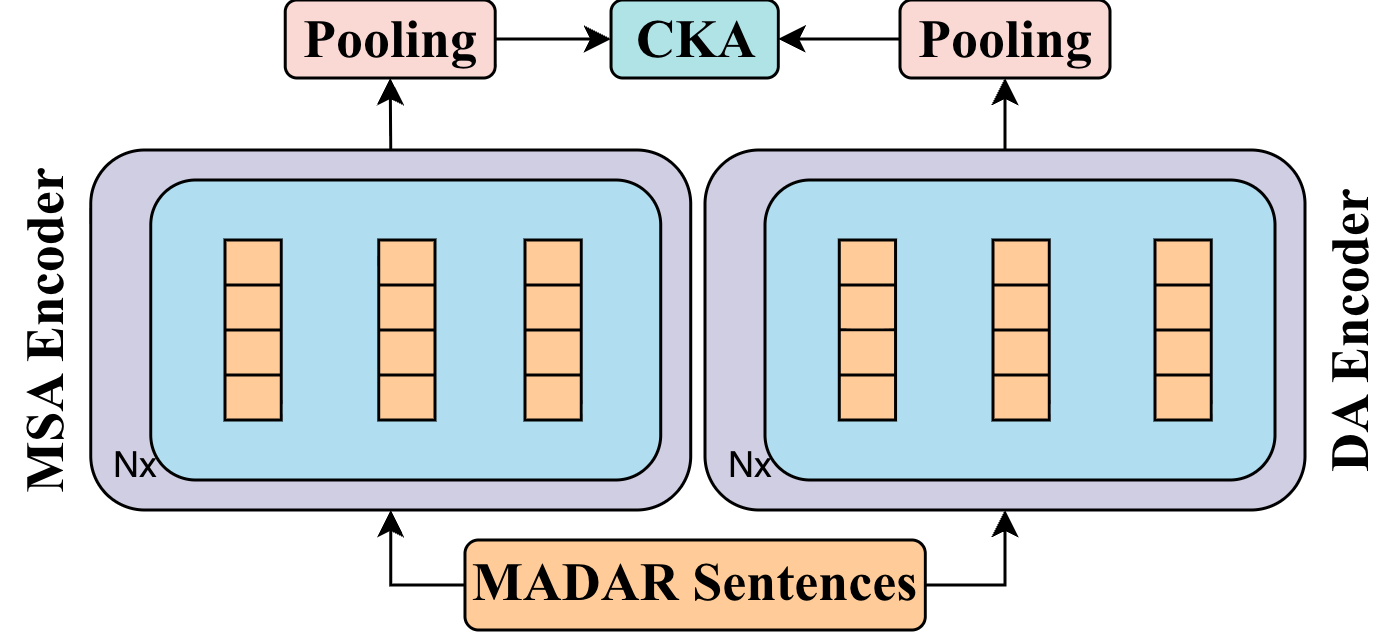}
    \caption{Architecture of CKA for representation similarity. MADAR parallel sentences are encoded by MSA and DA encoders through N layers, and the resulting representations are compared using linear CKA (Eq.~\ref{eq:cka_linear}).}
    \label{fig:CKA_diagram}
\end{figure}

Figure \ref{fig:CKA_diagram} illustrates our architecture for representation similarity, where we use Centered Kernel Alignment (CKA) \cite{pmlr-v97-kornblith19a} to measure the similarity of hidden representations. Let $X \in \mathbb{R}^{n \times d_x}$ and $Y \in \mathbb{R}^{n \times d_y}$ be representations of the same $n$ inputs. For linear CKA, we set $K = XX^\top$ and $L = YY^\top$ and compute:

\begin{equation}
\label{eq:cka_linear}
\small
\mathrm{CKA}(X, Y)
=
\frac{\lVert X^\top Y \rVert_F^2}
{\lVert X^\top X \rVert_F \,\lVert Y^\top Y \rVert_F}
\in [0, 1],
\end{equation}
where higher values indicate the representations are similar. Linear CKA is invariant to orthogonal transformations and isotropic rescaling of features, making it robust to rotations and global scaling differences between models.

This yields a compact and task-agnostic view of how similarly MSA and dialectal models encode Arabic input, complementing our probing-based analysis of functional transfer.

\subsection{Evaluating Dialectal Continuum}
\label{sec:geo_continuum}

We test whether similarity measures track geographical proximity continuum, i.e., whether models associated with countries closer (geographically) to a chosen MSA anchor exhibit higher similarity to an MSA-pretrained encoder. Similar approaches have been used for finding a relationship between language distance and cross-lingual transfer \cite{ERONEN2023103250, philippy-etal-2023-identifying}. A key challenge is that MSA has no fixed geographic locus, as it is used in formal contexts across all Arab countries. The literature proposes several proxies: one strand argues that MSA is particularly close to Palestinian (Levantine) Arabic, based on lexical similarity within Levantine and North-African varieties \cite{KWAIK20182}, however, this study did not include Gulf dialects. Other work highlights the strong affinity between Classical Arabic (CA) and Yemeni Arabic, citing the preservation of archaic forms, phonological features, and relative isolation, and notes that Gulf Arabic, despite borrowings (e.g., from Persian), maintains close ties to Bedouin dialects \cite{boyi2024relevance}. A separate lexical / lexical-semantic comparison with CA across regions also identifies Yemeni Arabic as the closest modern variety \cite{abouzahr2025lexical}; although the authors attribute this in part to corpus-size bias for Yemeni data, they also show that Gulf Arabic—geographically close to Yemen—has high similarity to CA. Finally, analysis of the MADAR corpus reports that Muscat (Oman) is the dialect most similar to MSA \cite{bouamor-etal-2018-madar}, and Muscat is likewise geographically near Yemen. Given the widely attested closeness of MSA to CA, we therefore \textbf{use Yemen as a geographic proxy for the MSA anchor}, while explicitly acknowledging this as an operational choice.

\section{Experimental Setup}
\subsection{Probing Datasets}

For fair cross-dialect comparison, we select dialectal datasets per task and enforce equal sample sizes per probe: for \textbf{POS}/\textbf{NER} we balance by number of labeled tokens, and for \textbf{SA} by number of sentences. General-purpose Arabic models are consistently evaluated against dialect-specific models on the latter’s native test sets. Table~\ref{dataset_list} details the corpora utilized for each dialect; for multi-dialectal datasets, we partition and reuse relevant subsets for their corresponding varieties. Due to the absence of publicly available NER corpora for Gulf dialects (specifically Riyadh and Muscat), we derive a silver-standard dataset from the MADAR corpus \cite{bouamor-etal-2018-madar}. We extract 2,000 sentences per dialect and perform automated entity annotation using the \textbf{CAMeL-Lab/bert-base-arabic-camelbert-da-ner} model \cite{inoue2021interplayvariantsizetask}.

\ifPDFTeX
    For the POS tagging task, we apply preprocessing tailored to Arabic cliticization, where multiple grammatical units are written as a single orthographic word. For example, \AR{وموبايل} (“\textbf{and mobile}”) combines \AR{و} (\emph{CONJ}) and \AR{موبايل} (\emph{NOUN}); we normalize such cases by aggregating clitic and stem tags into composite labels (e.g., \emph{CONJ+NOUN}), following prior work in Arabic POS tagging \cite{DARWISH18.562,habash-rambow-2006-magead}. NER datasets are used with their original tags and no additional preprocessing. For sentiment analysis, where data often comes from tweets and other social media, we remove URLs, HTML tags, usernames, and strip emojis using the \texttt{emoji}\footnote{\url{https://pypi.org/project/emoji/}} library.
\else
    For the POS tagging task, we apply preprocessing tailored to Arabic cliticization, where multiple grammatical units are written as a single orthographic word. For example, \foreignlanguage{arabic}{وموبايل} (“\textbf{and mobile}”) combines \foreignlanguage{arabic}{و} (\emph{CONJ}) and \foreignlanguage{arabic}{موبايل} (\emph{NOUN}); we normalize such cases by aggregating clitic and stem tags into composite labels (e.g., \emph{CONJ+NOUN}), following prior work in Arabic POS tagging \cite{DARWISH18.562,habash-rambow-2006-magead}. NER datasets are used with their original tags and no additional preprocessing. For sentiment analysis, where data often comes from tweets and other social media, we remove URLs, HTML tags, usernames, and strip emojis using the \texttt{emoji}\footnote{\url{https://pypi.org/project/emoji/}} library.
\fi

\paragraph{Dialect identification filtering.}
Because many probing datasets originate from noisy online sources, they may contain \emph{off-target} sentences (other dialects or MSA). To mitigate this, we run automatic dialect identification on every sentence using \textbf{CAMeL-Lab/bert-base-arabic-camelbert-mix-did-madar-corpus26} \cite{inoue2021interplayvariantsizetask}. City-level predictions (e.g., \emph{Riyadh}, \emph{Jeddah}, \emph{Cairo}) are mapped to countries, and we retain only sentences whose predicted country matches the dataset’s target dialect (or reassign them when the dataset explicitly spans multiple countries). This filtering step yields cleaner, better-aligned inputs for subsequent probing analyses.

\subsection{Similarity Analysis Dataset}
Our similarity analysis requires access to parallel sentences between MSA and dialects. Therefore, we use the \textbf{MADAR} corpus \cite{bouamor-etal-2018-madar}, which provides aligned versions of the same sentences in MSA and 25 city-level dialects. The corpus contains 2000 parallel sentences per city which helps avoid sample-size artifacts. 

\subsection{Models}

We focus on BERT-style encoders throughout this study, so that cross-model comparisons are not confounded by architectural differences and transfer effects remain directly comparable. Table~\ref{model_list} summarizes the models used and their key properties that help contextualize their behavior and performance across our experiments.

\begin{table*}[t]
\centering
\small 
\begin{tabular}{lllc}
\toprule
\textbf{Model Family} & \textbf{Model Name} & \textbf{Training Varieties} & \textbf{Tokens} \\ 
\midrule
\textbf{CAMeLBERT} & CAMeLBERT-MSA \cite{inoue2021interplayvariantsizetask} & MSA & 12.6B \\
 & CAMeLBERT-DA \cite{inoue2021interplayvariantsizetask} & DA & 5.8B \\
 & CAMeLBERT-MIX \cite{inoue2021interplayvariantsizetask} & MSA + DA + CA & 17.3B \\
\midrule
\textbf{Dialect-Specific} & SaudiBERT \cite{qarah2024saudibertlargelanguagemodel} & Saudi Dialect & 2.7B \\
 & EgyBERT \cite{qarah2024egybertlargelanguagemodel} & Egyptian Dialect & 1.07B \\
 & DarijaBERT \cite{gaanoun2023darijabert} & Moroccan Dialect (Darija) & 100M \\
 & DziriBERT \cite{abdaoui2022dziribertpretrainedlanguagemodel} & Algerian Dialect & 20M \\
\midrule
\textbf{AraRoBERTa} & AraRoBERTa-SA \cite{alyami-al-zaidy-2022-weakly} & Saudi Dialect & 45.4M \\
 & AraRoBERTa-EGY \cite{alyami-al-zaidy-2022-weakly} & Egyptian Dialect & 37.2M \\
 & AraRoBERTa-OM \cite{alyami-al-zaidy-2022-weakly} & Omani Dialect & 3.8M \\
 & AraRoBERTa-LB \cite{alyami-al-zaidy-2022-weakly} & Lebanese Dialect & 3.6M \\
 & AraRoBERTa-JO \cite{alyami-al-zaidy-2022-weakly} & Jordanian Dialect & 2.6M \\
 & AraRoBERTa-DZ \cite{alyami-al-zaidy-2022-weakly} & Algerian Dialect & 1.9M \\
\bottomrule
\end{tabular}
\caption{Overview of Arabic Pre-trained Language Models (PLMs)}
\label{model_list}
\end{table*}

\subsection{Probing} \label{functional_similarity}
\paragraph{Probe Training}
We employ a linear statistical classifier—multinomial logistic regression (softmax over a single affine layer)—implemented with the \textsc{NeuroX} toolkit \cite{dalvi2019neurox}. For the word-level tasks (POS and NER), we aggregate subword representations by selecting the embedding of the final subword token \cite{liu-etal-2019-linguistic}. For sentence-level Sentiment Analysis, representations are derived via mean-pooling across all tokens in the sequence.. We use stratified 80/20 train/test splits, the Adam optimizer with a small learning rate. Each layer $\ell$ is probed independently, producing a performance curve per model and task. Token-level embeddings for POS/NER are obtained by aggregating subword representations to the word level (using the embedding of the last subword token \cite{liu-etal-2019-linguistic}), while sentence-level embeddings for sentiment are computed by mean-pooling token representations over the sequence.    

\paragraph{Probe Evaluation}
We use the F1 score for the evaluation. We measure the performance of General models (i.e. MSA, Multi-DA and MIX) and DA models on both MSA only and DA datasets. We think that this offers a bidirectional look at the cross-lingual transfer of Arabic models. Where DA model that can transfer to MSA might indicate that the dialect is close to MSA and vice versa. Unless specified differently, we report the best F1 score across the model layers.

\subsection{Representation Similarity} \label{structural_similarity}
We use CKA to measure the representation similarity between MSA and DA models. We compute hidden states on the MADAR dataset, take the mean over the sentence tokens, then measure the layer-wise CKA similarity. The fact that MADAR is a parallel dataset allows us to investigate the following three scenarios:
\begin{enumerate}
    \item MSA model vs.\ Dialect model representation of DA sentences.
    \item MSA model vs.\ Dialect model representation of MSA sentences.
    \item MSA model vs.\ Dialect model representation of MSA and DA sentences respectively.
\end{enumerate}

Finally, since MADAR contains city level dialectal sentences and some countries are represented by more than one city (e.g., Egypt: Alexandria, Aswan, Cairo), we take the mean of CKA similarity over the cities to obtain the similarity per country, and consequently, per dialect.

\section{Results and Analysis}

\subsection{Probing Analysis}
\label{sec:results_probing}
\begin{figure}[tbp]
    \centering
    \includegraphics[width=0.48\textwidth]{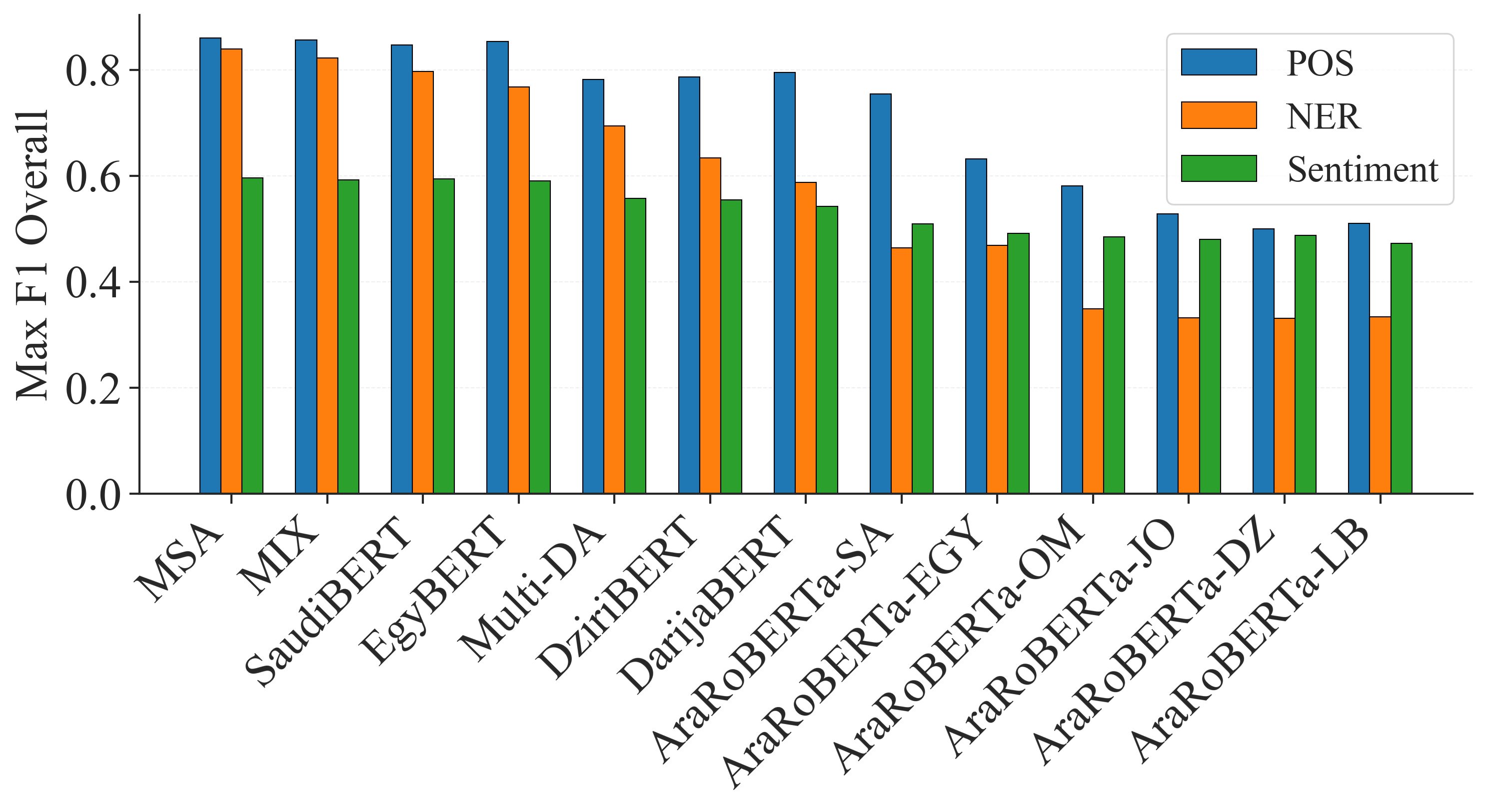}
    \caption{Performance of best performing layer on MSA Tasks.}
    \label{fig:msa_probing}
\end{figure}
For probing experiments implemented in Section \ref{functional_similarity}, we analyze and compare the performance of the best layer for each model. This is based on evidence that models store different information disproportionately in different layers \cite{tenney-etal-2019-bert}, while we are interested in comparing models in terms of their absolute performance rather than per-layer comparison.

\begin{figure}[tbp]
    \centering
    \includegraphics[width=0.48\textwidth]{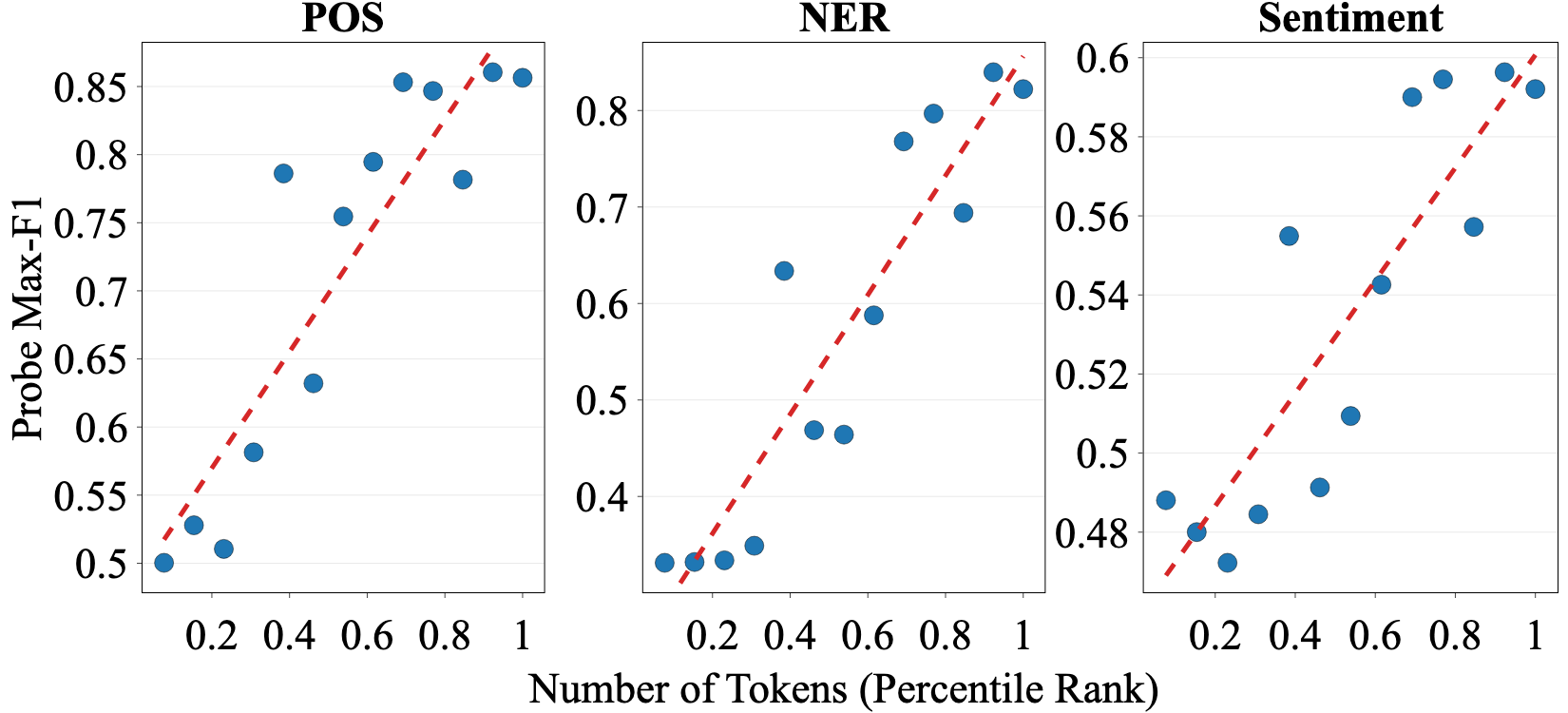}
    \caption{Impact of pretraining corpus size on probe performance across tasks. The Percentile Rank of the number of tokens is displayed for better visual interpretation.}
    \label{fig:Training_Size}
\end{figure}

Figure~\ref{fig:msa_probing} reports the macro-F1 of the best-performing layer for each model and task on the MSA datasets, as discussed in Section~\ref{functional_similarity}. Across POS, NER, and SA, a consistent pattern emerges: the \textbf{MSA-pretrained} model achieves the strongest performance on MSA text, outperforming both the \textbf{MIX} and \textbf{Multi-DA} models. Among dialect-specific encoders, \textbf{Saudi} and \textbf{Egyptian} models typically rank closest to the MSA model, while other regional models lag more substantially. The same trend holds within the \texttt{AraRoBERTa} family, where Saudi and Egyptian variants systematically outperform other dialectal variants on MSA inputs. By contrast, models trained on North African varieties, such as \texttt{DziriBERT} and \texttt{DarijaBERT}, generalize poorly to MSA across all tasks, likely reflecting the pronounced lexical and morphological divergence between Maghrebi dialects and MSA. When we relate model training size, as summarized in Table~\ref{model_list}, to performance on MSA data, we observe the trend visualized in Figure~\ref{fig:Training_Size}: models trained on larger and more diverse corpora tend to exhibit stronger generalization, underscoring the importance of corpus scale and linguistic variety for cross-dialectal robustness.

\begin{figure}[tbp]
    \centering
    \includegraphics[width=0.48\textwidth]{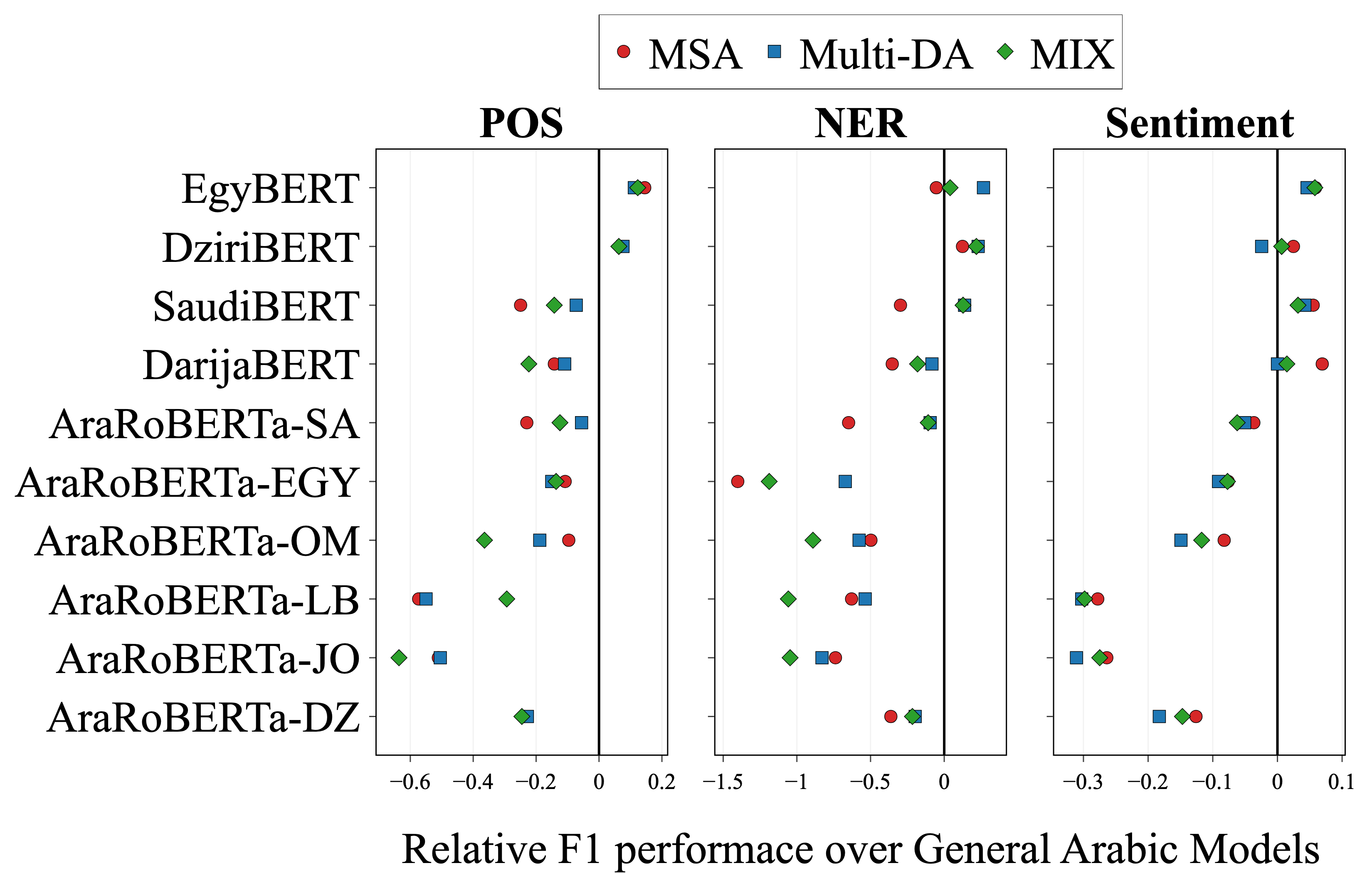}
    \caption{Relative performance of general vs. dialect-specific Arabic models on native dialectal datasets. Points to the right of the reference line denote cases where the dialect-specific model achieves higher performance, while points to the left indicate that the general model remains superior.\protect\footnotemark}
    \label{fig:dialectal_dataset_performance}
\end{figure}
\footnotetext{Non-normalized F1 score results are provided in our code repository.}

Figure \ref{fig:dialectal_dataset_performance} compares an \textbf{MSA} model, a \textbf{Multi-DA} model, and a \textbf{MIX} model against DA encoders on dialectal POS, NER, and SA. We compare models at their best-performing layer to control for layer specialization effects. We report the macro-F1 score performance of each model relative to the dialect specific model. We compute the relative difference as follows:

\begin{equation}
    \Delta = \frac{F1_{DA} - F1_M}{F1_{DA}}
\end{equation}
where $M = \{\text{MSA, Multi-DA, MIX}\}$. We normalize by $F1_{DA}$ to make the results comparable across dialects.

Overall, DA models such as \texttt{EgyBERT} and \texttt{DziriBERT} consistently excel on their native dialects—especially for SA and for POS where morphology and cliticization diverge most from MSA—highlighting the benefits of dialect-specific pretraining. The MSA model, while weaker on SA for most dialects, remains competitive on structural tasks (POS, NER) for several varieties, indicating that MSA pretraining preserves transferable syntactic and entity-structure regularities. The Multi-DA model provides a strong dialect-agnostic baseline, where it often outperforms the MSA model on the SA task, but often falls short of the strongest DA models on their home dialect although it is trained on more data. This suggests that training on multiple Arabic dialects can still lead to negative interference \cite{wang-etal-2020-negative}, although they are regarded as similar in literature. The MIX model shows a similar breadth–sharpness trade-off: it can match or surpass many dialectal models on POS, but tends to underperform DA encoders on NER and SA when domain and style are highly dialect-specific. Across these comparisons, pretraining scale emerges as a key factor. DA models that reliably beat the general-purpose encoders (MSA, Multi-DA, MIX) are often trained on substantially larger pretraining corpora than lighter dialectal models (e.g., \texttt{EgyBERT} vs.\ \texttt{AraRoBERTa-EGY}, \texttt{DziriBERT} vs.\ \texttt{AraRoBERTa-DZ}). This suggests that the amount of pretraining data is crucial to improve the model's encoding of dialect-specific morphological, syntactic, and semantic features.

\subsection{Representation Similarity Analysis}
\label{sec:results_cka}
\begin{figure}[tbp]
    \centering
    \includegraphics[width=0.48\textwidth]{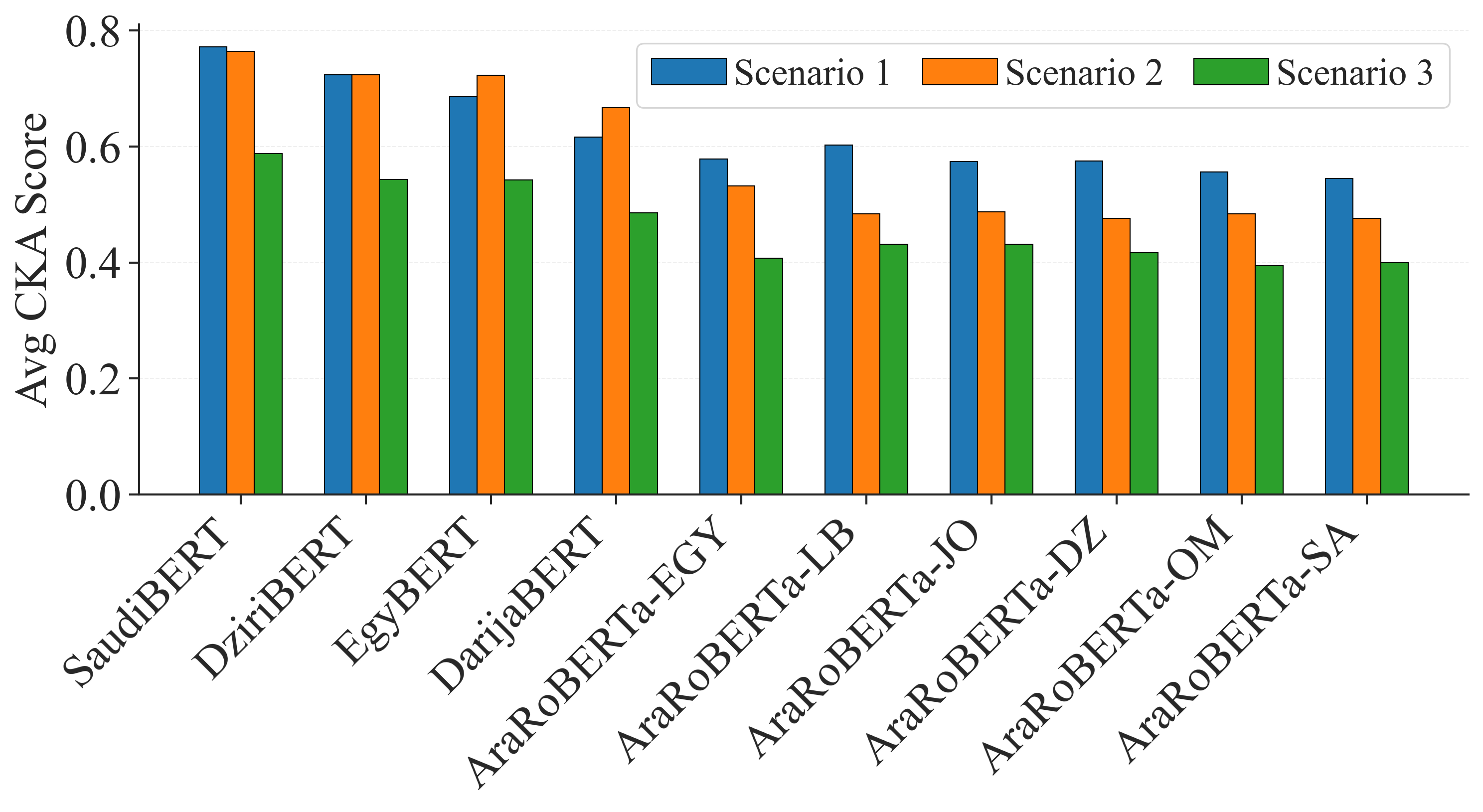}
    \caption{Average layer-wise CKA between the MSA model and DA models for the three scenarios described in Section \ref{structural_similarity}.}
    \label{fig:avg_cka}
\end{figure}

Figure~\ref{fig:avg_cka} summarizes the average layer-wise CKA between the MSA model and each DA model across all three scenarios discussed in Section \ref{structural_similarity}. When both models encode the same dialectal sentences (Scenario 1), the highest alignment is observed for \texttt{SaudiBERT} on Saudi text, followed by \texttt{DziriBERT}, \texttt{EgyBERT}, and \texttt{DarijaBERT}, with \texttt{AraRoBERTa-LB} showing the closest alignment among the \texttt{AraRoBERTa} variants. When both encoders process the MSA versions of the same sentences (Scenario 2), the order of BERT-based dialect models remains similar, while within the \texttt{AraRoBERTa} family, \texttt{AraRoBERTa-EGY} becomes the closest to the MSA encoder and the remaining variants form a slightly lower, tightly clustered group. In scenario 3, where the MSA model encodes MSA sentences and dialectal models encode parallel dialect sentences, the BERT models ranking largely persists (\texttt{SaudiBERT} $>$ \texttt{DziriBERT} $>$ \texttt{EgyBERT} $>$ \texttt{DarijaBERT}), whereas within \texttt{AraRoBERTa}, Levantine variants (\texttt{AraRoBERTa-JO}, \texttt{AraRoBERTa-LB}) show the strongest alignment, followed by North-African variants (\texttt{AraRoBERTa-DZ}, \texttt{AraRoBERTa-EGY}), then Gulf variants (\texttt{AraRoBERTa-SA}, \texttt{AraRoBERTa-OM}). Across scenarios, scenario 3 has the lowest similarity scores, showing the representations of DA and MSA models are not agnostic to their input variant. Furthermore, even when the input is the same variant (i.e. scenario 1 and 2), the similarity is always lower than 0.8, showing that MSA models fail to capture DA specific nuances and vice versa. The results of AraRoBERTa models suggest that this can be partly explained by the amount of training data, where the similarity of these models is lower than the BERT-based models, which are trained on more data.

\subsection{Proximity Analysis}

\begin{figure*}[tbp]
    \centering
    \begin{subfigure}[b]{\columnwidth}
        \centering
        \includegraphics[width=\textwidth]{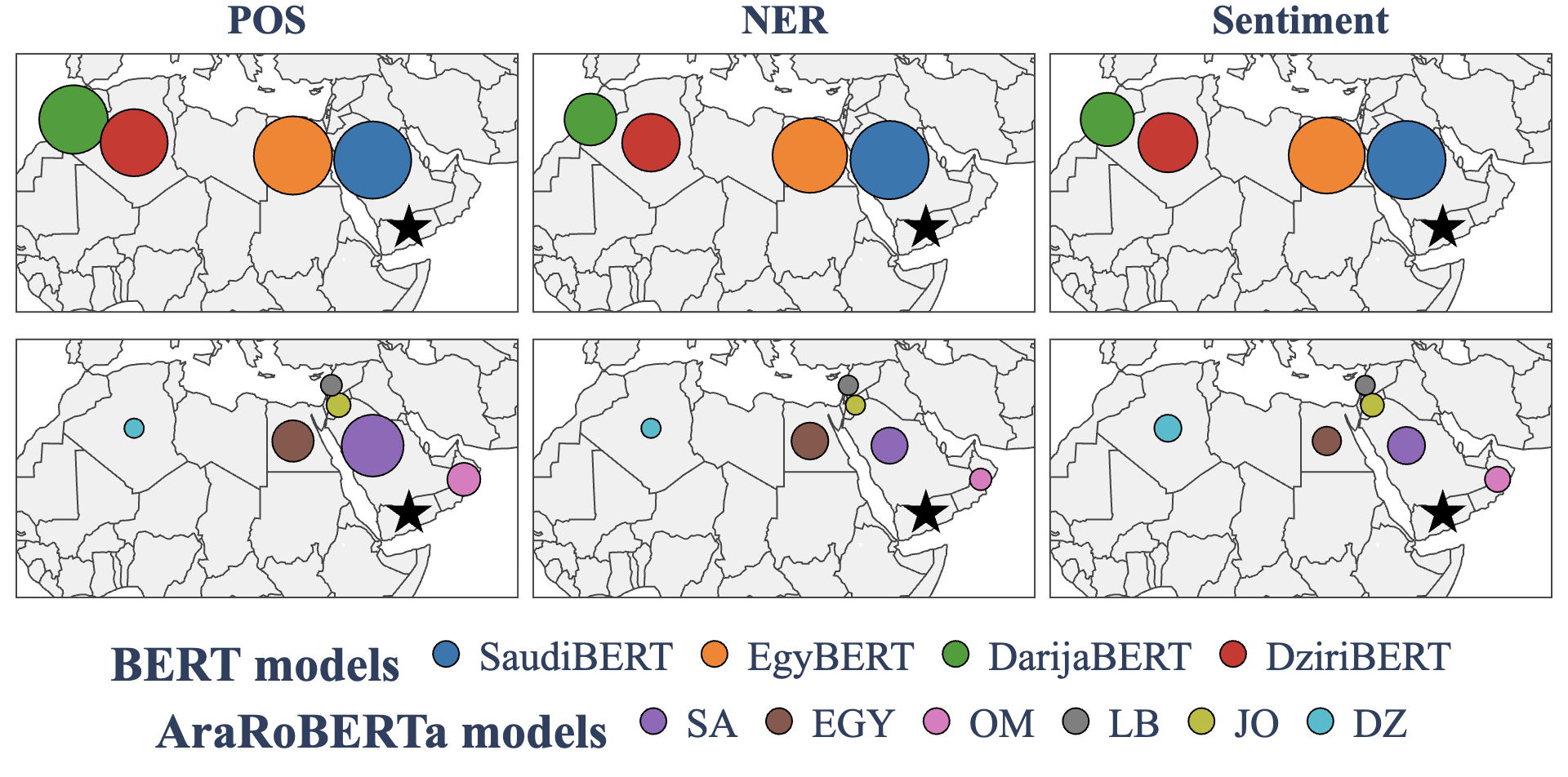}
        \caption{Geographic proximity vs.\ probing performance.}
        \label{fig:Probing_Continuum}
    \end{subfigure}
    \hfill 
    \begin{subfigure}[b]{\columnwidth}
        \centering
        \includegraphics[width=\textwidth]{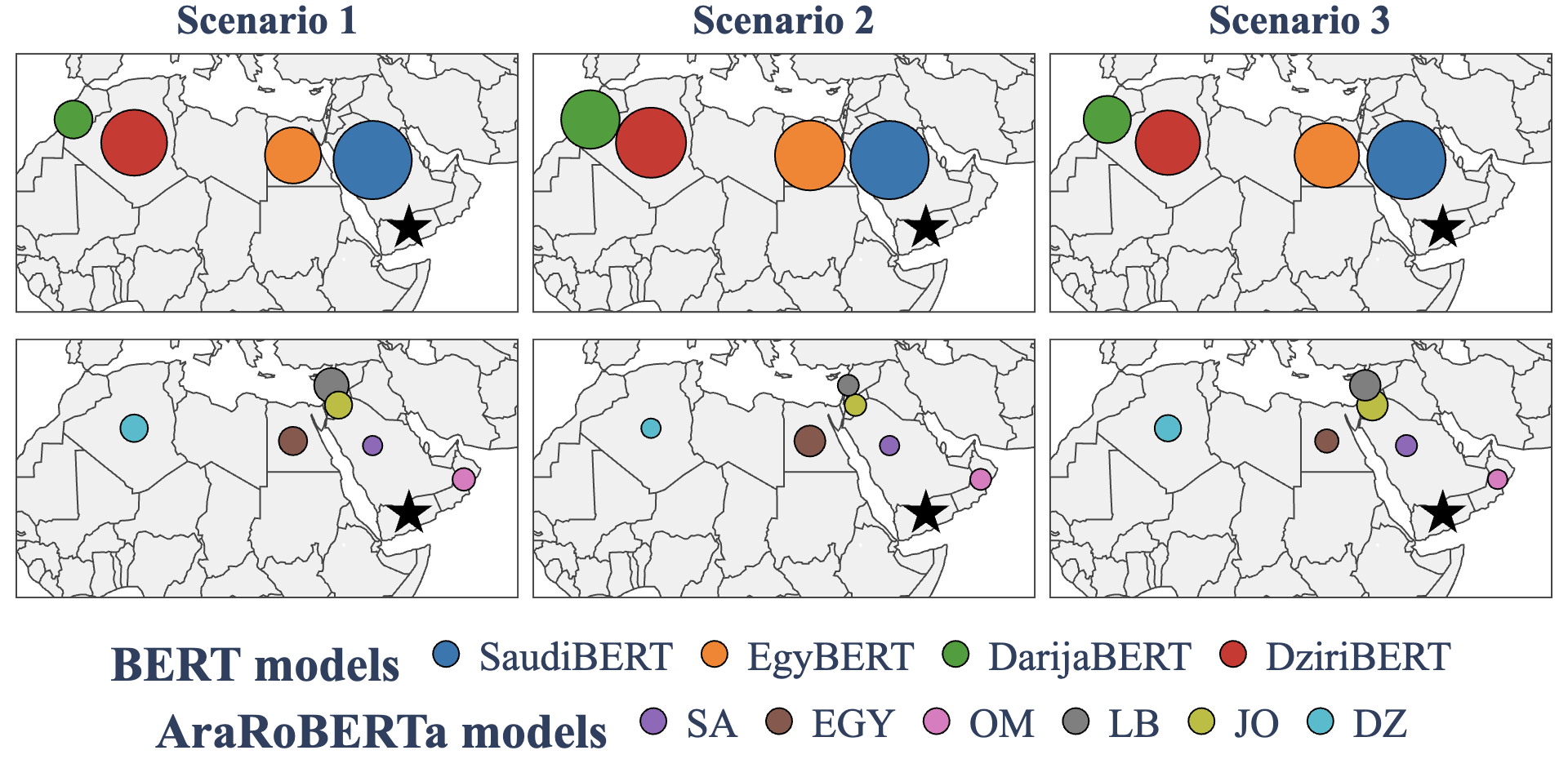}
        \caption{Geographic proximity vs.\ CKA-based similarity.}
        \label{fig:CKA_Continuum}
    \end{subfigure}

    \caption{Comparison of Geographic Proximity with Probing Performance and CKA Similarity. The black star marks Yemen (the MSA anchor), and each bubble denotes a DA model; bubble size encodes (a) macro-F1 score on MSA and (b) average CKA score.}
    \label{fig:combined_continuum}
\end{figure*}

As discussed in Sections \ref{sec:results_cka} and \ref{sec:results_probing}, there is significant disparity between dialect specific models in terms of their probing performance and similarity with MSA models. Motivated by the documented linguistic phenomena of dialect continuum \cite{Chambers_Trudgill_1998}, and by previous studies that investigated the relationship between geographic proximity and lexical overlap \cite{ALSUDAIS2022102770}, we study whether geographic proximity can explain the disparity in probing and representation similarity results. Note that we use Yemen as the geographic location for MSA as explained in Section \ref{sec:geo_continuum}. Figure~\ref{fig:Probing_Continuum} shows dialectal models by their geographic location, with bubble sizes proportional to their macro-F1 score on MSA text from the probing tasks. Across all tasks, we observe a clear dialectal continuum: models associated with regions geographically closer to the MSA anchor are represented by larger bubbles, which gradually shrink with increasing distance. This pattern indicates that cross-lingual transfer in terms of \emph{functional} similarity aligns well with geographic proximity. Figure~\ref{fig:CKA_Continuum} presents an analogous view for CKA-based similarity. Here, the dialectal continuum is evident primarily for models trained on large-scale corpora, while models with more limited pretraining data (i.e. \texttt{AraRoBERTa} models) do not reflect a dialect continuum effect. The results suggest that proximity plays a role in model performance and similarity, and is in line with the phenomena of dialect continuum, where more distant dialects and less mutually intelligible, which leads to weaker cross-lingual transfer and lower similarity. Specifically, Scenario 2 in Figure~\ref{fig:CKA_Continuum} is more comparable with Figure~\ref{fig:Probing_Continuum} since we compare the dialectal continuum of the MSA model and the DA models on MSA text.

\section{Discussion}

Based on the probing results on MSA datasets, we observe that the MSA model outperforms both MIX and Multi-DA models. The relative underperformance of the MIX model aligns with prior findings that monolingual MSA encoders surpass mixed-variant models on MSA tasks \cite{inoue2021interplayvariantsizetask}. Although the MIX model is trained on MSA data, the underperformance can be due to curse of multilinguality which adversely affects high-resource varieties \cite{chang2023multilingualitycurselanguagemodeling}. Among the DA models we observe that cross-lingual transfer to MSA is strongly dependent on the amount of data the model is trained on. This observation is further validated by geographical proximity analysis where models trained on large amount of data capture dialect continuum more efficiently. We also observe that it is easier for the language models to capture functional similarity even with less training data. But it is more difficult for Arabic language models to capture structural similarity as observed from geographical proximity analysis. Furthermore, probing and CKA reveal a clear functional–structural gap: some models (e.g., \texttt{DziriBERT}) can be structurally close to MSA yet fail to convert this into superior probing performance. This is in line with prior work showing that high CKA similarity does not guarantee functionally similar or equally useful features \cite{davari2022reliabilityckasimilaritymeasure}. 

The probing performance on the DA datasets shows that the MSA-pretrained encoder is generally stronger on NER and POS tasks, while DA models tend to excel on SA tasks. The performance on POS suggests that Arabic and its dialects maintain a similar syntax, while the performance on NER can be attributed to code switching, where DA varieties often use MSA for entities. However, this does not translate to a higher performance on SA, where higher level semantic and pragmatic understanding is required to predict the sentiment. These findings suggest that Arabic dialects are similar to MSA in their syntax and use code switching for entities which explains their performance on POS and NER tasks respectively, however, they struggle on SA which is characterized by dialect specific idioms, non-standard orthography and distinct negation patterns that are absent from their pretraining data.

Comparing DA models with \emph{general} dialectal models (Multi-DA; MIX) on downstream tasks (Figures~\ref{fig:dialectal_dataset_performance}) shows that \emph{general} models frequently achieve stronger average performance across dialects, a likely consequence of larger and more diverse pretraining. The MIX model commonly surpasses the Multi-DA model, suggesting that incorporating MSA and Classical Arabic alongside dialectal text improves the encoding of morphological and syntactic regularities. Nonetheless, there are notable exceptions: \texttt{DziriBERT} and \texttt{EgyBERT} often outperform \emph{general} models on their native dialects. For SA, multiple DA models (with the exception of some \texttt{AraRoBERTa} variants) outperform general models, while the MIX model retains advantages on NER and POS tasks, where training on MSA text gives an advantage. We see that for most high resource dialects (Egyptian, Saudi Arabia), DA models usually outperform \emph{general} dialect models. Whereas for low resource dialects (Lebanese, Jordan), \emph{general} dialect models outperform native dialectal models. This observation is consistent with recent research on curse of multiliguality \cite{chang2023multilingualitycurselanguagemodeling,wu-dredze-2020-languages}, where multi-dialect pretraining benefits low resource dialects, but negatively affects high resource dialect performance.

\section{Conclusion} \label{conclusion}

In this work, we study how Arabic language models (LMs) capture cross-lingual transfer among dialects and their relationship to Modern Standard Arabic (MSA), combining \textbf{functional} evidence from probing classifiers (POS, NER, SA) and \textbf{structural} evidence from representation similarity (CKA), and linking both to \textbf{geographic proximity}. Our analysis covered MSA, MIX, and Multi-DA encoders alongside mono-dialectal DA models. We find consistent signals of a dialectal continuum: models associated with geographically closer varieties tend to align more closely with MSA in both probing and CKA, but this effect strongly depends on the amount of pretraining data. Interestingly, even among dialects of the same language we find evidence for negative interference, where Multi-DA models underperform mono-dialectal models of high resource dialects. Our findings question the assumption of cross-lingual transfer from MSA to its dialects, where transfer is feasible but disproportionate across varieties. Furthermore, Multi-DA models might benefit from dialect or dialect-region specific parameters to avoid negative interference.

\section*{Limitations} \label{limitations}

Our work has the following limitations:

\noindent
\textbf{Dialectal dataset coverage:} No publicly available Gulf-dialect NER dataset exists to our knowledge; we therefore leveraged \textsc{MADAR} and produced proxy annotations using an existing NER model. This introduces potential annotation noise and domain mismatch that may affect NER-related comparisons. Additionally, we rely on regional datasets (e.g., Gulf, Maghrebi) that are further segmented by country (e.g., Saudi, Moroccan) using automatic dialect identification. However, dialect identification is imperfect, and occasional false positives may introduce noise into the country-level splits and, consequently, into our downstream analyses.

\noindent
\textbf{Geographic proxy for MSA:} Our choice of Yemen as a geographic proxy for MSA is literature-informed but not definitive. Results involving distance-to-anchor should be interpreted as operational rather than canonical; alternative anchors could yield different effect sizes.

\noindent
\textbf{Model heterogeneity:} The compared models differ in training size, vocabularies, and pretraining hyperparameters. Such heterogeneity can confound attribution of observed effects. A more controlled comparison—equalizing corpus size, and training settings—would yield more robust findings.

\noindent
\textbf{Pretraining data composition:} The paper does not focus the effect of the composition of pre-training data of dialectal models. Some models, such as \texttt{DarijaBERT}, were trained by filtering out MSA data, where as \texttt{SaudiBERT}, \texttt{DziriBERT}, and \texttt{EgyBERT} were trained on data collected based on geolocation, which may include traces of MSA data. The presence of noisy MSA data in dialectal models may affect the model's performance.


\bibliography{bibliography_paper}

\appendix

\section{Probing Datasets}
Table \ref{dataset_list} presents the datasets we use for probing, where we specify the dialects they support and the dialects we used them for.

\begin{table*}[]
\centering
\resizebox{\textwidth}{!}{%
\begin{tabular}{llllll}
\hline
\textbf{Task} & \textbf{Dataset} & \textbf{Dataset Dialect} & \textbf{Used for Dialect} & \textbf{Total Samples} & \textbf{Dialect Samples} \\ \hline
\multirow{5}{*}{POS} & NArabizi \cite{seddah-etal-2020-building} & Algeria & Algeria & 1279 & 731 \\
 & QCRI Arabic POS Dialect \cite{DARWISH18.562} & Egypt, Levant, Gulf, Maghreb & Egypt, Morocco & 350, 350, 350, 350 & 338, 145 \\
 & Shami \cite{abu-kwaik-etal-2018-shami} & Levant & Lebanon, Jordan & 1069 & 240, 159 \\
 & CAMeL Treebank \cite{habash-etal-2022-camel} & MSA and CA & MSA & 5755 (MSA Variant, 21st Centuary) & 2399 \\
 & GUMAR \cite{khalifa-etal-2016-large} & Gulf & Saudi, Oman & 15205 & 1774, 1633 \\ \hline
\multirow{6}{*}{NER} & CLEANANERcorp \cite{alduwais-etal-2024-cleananercorp} & MSA & MSA & 4899 & 3050 \\
 & ACDNER \cite{elkhbir-etal-2023-cross} & Egypt, Morocco, Syria & Egypt & 353, 378, 361 & 210 \\
 & DzNER \cite{DAHOU2023100005} & Algeria & Algeria & 5859 & 1153 \\
 & DarNERcorp \cite{MOUSSA2023109234} & Morocco & Morocco & 2511 & 2177 \\
 & Wojood \cite{jarrar-etal-2022-wojood} & MSA, Lebanon, Jordan & Lebanon, Jordan & 24643, 383, 4705 & 190, 570 \\
 & MADAR (Tagged) \cite{bouamor-etal-2018-madar} & Multi-Dialect, MSA & Saudi, Oman & 2000 each & 2000, 2000 \\ \hline
\multirow{7}{*}{SA} & DZYT \cite{app132011157} & Algeria & Algeria & 49942 & 7828 \\
 & AET \cite{DVN/LBXV9O_2019} & Egypt & Egypt & 39993 & 24546 \\
 & JHSC \cite{10.3389/frai.2024.1345445} & Jordan & Jordan & 276003 & 30603 \\
 & L-HSAB \cite{mulki-etal-2019-l} & Lebanon & Lebanon & 5377 & 1142 \\
 & LABR \cite{aly-atiya-2013-labr} & MSA & MSA & 63257 & 21696 \\
 & MYC \cite{Jbel_2024} & Morocco & Morocco & 19455 & 6306 \\
 & MARSA \cite{9576756} & Gulf & Saudi, Oman & 68128 & 10146, 2906 \\ \hline
\end{tabular}%
}
\caption{Overview of Probing Datasets}
\label{dataset_list}
\end{table*}

\end{document}